\documentclass[letterpaper, 10 pt, conference]{ieeeconf}  

\IEEEoverridecommandlockouts                              

\usepackage{graphicx}
\usepackage{cite}

\title{\LARGE \bf
Automated Construction of Time-Space Diagrams for Traffic Analysis Using Street-View Video Sequences
}

\author{Tanay Rastogi$^{*,1}$ and Mårten Björkman$^{1}$
\thanks{$^{1}$ All authors are with KTH Royal Institute of Technology, Stockholm, Sweden; Email: {\tt\small tanay@kth.se, celle@kth.se}}%
\thanks{$^{*}$ Correspondence Email: tanay@kth.se}%
}

\begin{document}
\maketitle
\thispagestyle{empty}
\pagestyle{empty}

\begin{abstract}
Time-space diagrams are essential tools for analyzing traffic patterns and optimizing transportation infrastructure and traffic management strategies. Traditional data collection methods for these diagrams have limitations in terms of temporal and spatial coverage. Recent advancements in camera technology have overcome these limitations and provided extensive urban data. In this study, we propose an innovative approach to constructing time-space diagrams by utilizing street-view video sequences captured by cameras mounted on moving vehicles. Using the state-of-the-art YOLOv5, StrongSORT, and photogrammetry techniques for distance calculation, we can infer vehicle trajectories from the video data and generate time-space diagrams. To evaluate the effectiveness of our proposed method, we utilized datasets from the KITTI computer vision benchmark suite. The evaluation results demonstrate that our approach can generate trajectories from video data, although there are some errors that can be mitigated by improving the performance of the detector, tracker, and distance calculation components. In conclusion, the utilization of street-view video sequences captured by cameras mounted on moving vehicles, combined with state-of-the-art computer vision techniques, has immense potential for constructing comprehensive time-space diagrams. These diagrams offer valuable insights into traffic patterns and contribute to the design of transportation infrastructure and traffic management strategies.

\end{abstract}

\section{INTRODUCTION}
Time-space diagrams are graphical representations that illustrate vehicle trajectories over time in a specific road section. They provide valuable insights into microscopic traffic characteristics (e.g., time and space headways) as well as macroscopic characteristics such as density, flow, and mean speed. These diagrams play a crucial role in designing transport infrastructure, implementing traffic management strategies, improving safety, and reducing the environmental impact of transportation. Numerous studies have leveraged time-space diagrams to derive optimal approaches to traffic control \cite{Wang2020ATrajectories, Peng2023Network-WideTrajectories, Essa2018TrafficLevel}. In addition, various methods have been proposed to predict traffic congestion through the application of deep learning techniques and training networks using information derived from space-time trajectories \cite{Xing2022AFlow, KhajehHosseini2023TowardsNetwork}. Furthermore, time-space diagrams have been used to estimate traffic states \cite{Nantes2016Real-timeData, Thodi2021LearningVisualizations},  and offer valuable insights to formulate policy recommendations \cite{Zang2019ExperimentalBeijing, Li2017AnalysisArrivals}.

The primary data source utilized for generating time-space trajectories is commonly derived from Floating Car Data (FCD). FCD involves employing mobile sensors installed within vehicles, such as Global Positioning System (GPS) devices, On-board Diagnostics systems (OBD), or embedded cameras, to capture measurements during their respective trips. Numerous research studies have extensively used FCD data to estimate crucial traffic parameters such as flow, speed, and density \cite{Zhang2022MonocularReview, Sunderrajan2016TrafficData, Yuan2021TrafficData}. However, as emphasized in \cite{Rahmani2010RequirementsStudy}, any analysis based on FCD data requires prior knowledge of the penetration rate, which can be challenging to obtain. 

The advancement of camera technology has revolutionized the field of traffic data collection and analysis. Cameras function as sensors capable of capturing real-time data to obtain traffic parameters such as mean speed, density, and level of service for specific sections of the road network within their field of view. Extensive research has been conducted in traffic flow analysis utilizing high-mount surveillance or highway cameras \cite{Ua-Areemitr2019Low-CostProcessing,Pletzer2012RobustCameras}. The recent advancements in lightweight portable sensors, including smartphones, dash cams, and in-vehicle cameras, have expanded the range of possibilities for data collection. There has been a growing focus on data collection using moving cameras, particularly for Traffic State Estimation (TSE). For instance,  \cite{Seo2015EstimationEquipment, Cao2011MobileIntegration} have developed solutions using moving cameras to collect traffic data from vehicles in the same lane to calculate macroscopic traffic flow and speed. Furthermore, other studies \cite{Guerrieri2021DeepTechnique, Kumar2021CitywideVideos} have utilized on-board cameras to estimate TSE, with a specific emphasis on analyzing traffic in the opposite lane. These advances highlight the expanding utility of cameras and their application in traffic research and analysis.

This paper aims to introduce a novel approach to creating time-space diagrams for capturing traffic flow characteristics using an in-vehicle camera mounted on a moving vehicle. The proposed method employs state-of-the-art computer vision algorithms based on deep neural networks (DNN), photogrammetry, and geodesy to construct time-space diagrams from a video sequence. Unlike stationary sensors, the proposed method overcomes spatial limitations, and unlike mobile sensors, it does not rely on prior knowledge of the number of probe vehicles on the road. In contrast to previous research utilizing moving cameras, our focus is specifically on capturing microscopic traffic flow characteristics, distinguishing it from the aforementioned studies.

The remainder of the article is structured as follows: Section \ref{sec:methodology} describes the proposed traffic data collection approach, as well as the computer vision algorithms employed in the proposed algorithm and the dataset used to validate the proposed methodology. Section \ref{sec:results} presents the experimental results of the test on the proposed methodology as well as a discussion of the results. Finally, Section \ref{sec:conclusion} summarizes the findings of the experiment and concludes the article with suggestions for future work.

\section{Methodology}\label{sec:methodology}
\begin{figure}
\centering
\includegraphics[width=3.4in]{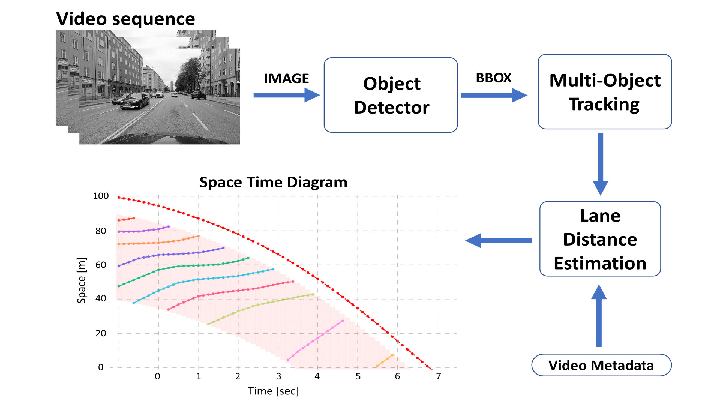}
\caption{Flow chart that illustrates the proposed methodology of extracting the vehicle trajectory from a street-view video sequence using different computer vision algorithms.}
\label{fig:pipeline}
\end{figure}

The proposed methodology aims to generate time-space diagrams utilizing trajectories inferred from street-view videos captured by an on-board camera mounted on a moving vehicle along a specific link on a road network. This methodology involves three key steps: multi-object detection, multi-object tracking, and estimation of lane distance, as shown in Fig. \ref{fig:pipeline}. For the detection and identification of vehicles in each frame of the video sequence, we employed the YOLOv5 multi-object detector. Subsequently, each detected vehicle in the image is assigned a unique ID and tracked across consecutive frames using a multi-object tracking method known as StrongSORT, which employs tracking-by-detection. Once vehicles have been detected and labeled, the distance of each detected vehicle link on a road network is calculated using time-stamped GPS information, photogrammetry, and geodesy. Finally, utilizing the distance and timestamp information, the time-space diagram is generated, with the spatial dimensions corresponding to the link length and the time dimension representing the travel time of the camera-mounted vehicle.


The methodology presented in this study is based on certain assumptions regarding the data collection process. Specifically, the methodology focuses on the \textit{car-like} object class and is tested accordingly. While the proposed method has the capability to detect and track multiple objects, such as pedestrians, it is important to note that these objects are deemed irrelevant for the intended application of time-space trajectories. The position and length of the link in the road network are considered predetermined information. When extracting vehicle trajectories from the street-view videos, only vehicles visible in the video and traveling in the opposite lane are considered. This ensures that distinct trajectories are obtained for all vehicles traversing the link. Vehicles in the same lane as the moving camera are not included, as their speeds are usually similar, resulting in less informative samples. This approach aligns with previous research that aimed to analyze traffic flow and speed by considering vehicles in the opposite lane \cite{Guerrieri2021DeepTechnique, Kumar2021CitywideVideos}. 

The suggested approach has solely been evaluated on real-world videos present within the dataset. To extend the evaluation of this approach concerning the impacts of traffic demand and congestion levels, simulated data is required. This simulated data would enable the creation of various traffic scenarios for testing, similar to the methodology employed in \cite{Kumar2022CARLA}.


\subsection{Dataset} \label{dataset}
The analysis in the article relies on an open-source computer vision benchmark suite called KITTI \cite{Geiger2012AreSuite}. KITTI comprises a comprehensive collection of high-resolution images, videos, and 3D point clouds, along with accurately calibrated camera and LiDAR data. The dataset was gathered by driving a car through urban environments, ensuring its relevance to real-world scenarios. 


We used annotated images from the \textit{KITTI 2D object detection} dataset to train and validate the proposed object detector. The 7,481 images in the dataset are labeled with bounding boxes and classifications for a total of eight categories, namely car, cyclist, misc, pedestrian, person-sitting, tram, truck, and van. The dataset is split into 70\%-20\%-10\% ratios for training, validation, and testing sets, respectively, in such a way that each set has the same proportion of labels in it. Notably, the dataset exhibits a significantly higher number of objects labeled "car" compared to other categories \cite{Geiger2012AreSuite}. This skewed distribution is appropriate for the purposes of this article, since the focus is primarily on the detection of cars.



To evaluate the performance of the multi-object tracker within the proposed methodology, we utilized annotated video sequences from the \textit{KITTI 2D box tracking} dataset. This dataset comprises 21 videos, each annotated with information such as tracking ID, object type, 2D/3D bounding box dimensions, and object location in camera coordinates. We used KITTI's evaluation suite, TrackEval \cite{Luiten2020TrackEval}, which allowed us to compare the proposed multi-object tracker's performance to that of other well-known trackers and evaluate the tracker's output on the annotated videos.

Furthermore, object location information from the same set of videos was utilized to analyze the performance of the proposed lane distance estimation method. This information provided valuable insights into estimating the distance of objects. Lastly, we used the raw GPS data for the videos in the \textit{KITTI 2D box tracking} dataset to generate ground truth time-space diagrams. Subsequently, these diagrams were compared to the time-space diagrams generated from the proposed methodology, providing a means to evaluate the proposed approach.

\subsection{Object Detection}\label{sec:veh_detection}
Within the proposed methodology, we employed the YOLOv5 object detector network, which is part of the You Only Look Once (YOLO) family of single-shot detectors. YOLO networks are known for their exceptional speed and accuracy \cite{Nepal2022ComparingUAVs., GeYOLOX:Detectors}. These networks make predictions based on predefined anchors, representing potential object locations in the image, with varying box and aspect ratios centered at each pixel.

In the proposed methodology, we utilized the YOLOv5 network, developed by Glen Jocher and maintained by an open-source community called Ultralytics \cite{GlennJocher2022YOLOv5}. This network was trained on the MS-COCO dataset \cite{Lin2014MicrosoftContext} and can detect a wide range of objects across eighty general categories, including classes such as car, bus, truck, and pedestrian. YOLOv5 has demonstrated state-of-the-art performance in object detection on the MS-COCO dataset \cite{GlennJocher2022YOLOv5}.


Specifically, we re-trained the YOLOv5m model on the KITTI 2D object detection dataset, tailoring it to our proposed methodology. During the training process, we froze the first 15 layers of the YOLOv5m model, utilizing weights from the pre-trained network. The remaining layers were trained using the KITTI 2D object detection training set, which consisted of 5233 annotated images, while 1495 images were reserved for the validation set. Table \ref{tbl:yolo_hyps} list the specific hyperparameters used for training the model. We retrained the YOLOv5m network on a GRID T4-8C GPU with 8GB memory for $200$ iterations. Subsequently, the model that achieved the highest mean Average Precision (mAP) on the validation set was chosen for further analysis and evaluation.

\begin{table}
\renewcommand{\arraystretch}{1.3}
\caption{Hyperparameters used for training YOLOv5 on KITTI dataset}
\label{tbl:yolo_hyps}
\centering
\begin{tabular}{|l|l|}
\hline
\multicolumn{1}{|c|}{\textbf{Hyperparmaters}} & \multicolumn{1}{c|}{\textbf{Value}} \\ \hline
Input shape                                   & (1280, 1280)                          \\ \hline
Batch size                                    & 16                                  \\ \hline
Epochs                                        & 200                                 \\ \hline
Learning rate                                 & 0.01                                \\ \hline
Activation function                           & SiLU\cite{Elfwing2018Sigmoid-weightedLearning}\\ \hline
Optimizer                                     & SGD with momentum (0.937)           \\ \hline
IoU threshold for training                    & 0.2                                 \\ \hline
\end{tabular}
\end{table}

\subsection{Multi-Object Tracking}\label{sec:veh_tracking}
To generate trajectories for the detected vehicles, it is necessary to track their movements in multiple frames in the video sequence. In the proposed methodology, we utilize a tracking-by-detection multi-object tracker called StrongSORT \cite{Du2022StrongSORT:Again}. This method builds on the classical DeepSORT tracker by introducing improvements in detection, embedding, and association.


In the proposed methodology, we utilize the PyTorch implementation of StrongSORT \cite{Brostrom2022RealtimeOSNet}. We replace the object detector in StrongSORT with the YOLOv5m network trained on the KITTI dataset, described in Section \ref{sec:veh_detection}. The remaining hyperparameters for the StrongSORT tracker are kept as defaults for the analysis, and they are listed in Table \ref{tbl:sort_hyps}.

\begin{table}
\renewcommand{\arraystretch}{1.3}
\caption{Hyperparameters used for StrongSORT tracker \cite{Brostrom2022RealtimeOSNet}}
\label{tbl:sort_hyps}
\centering
\begin{tabular}{|p{0.1\linewidth}|p{0.45\linewidth}|c|}
\hline
\multicolumn{1}{|c|}{\textbf{Hyperparmaters}} & \multicolumn{1}{c|}{\textbf{Explanation}}                 & \textbf{Value} \\ \hline
NN\_metric                                    & Nearest neighbour distance metric                         & cosine         \\ \hline
Max\_dist                                     & Nearest neighbour matching threshold                      & 0.2            \\ \hline
Max\_IoU\_dist                                & Max association threshold                                 & 0.7            \\ \hline
Max\_age                                      & Maximum number of missed misses before a track is deleted & 30             \\ \hline
Max\_init                                     & Number of frames before object gets tracked               & 2              \\ \hline
\end{tabular}
\end{table}

\subsection{Lane Distance Estimation}\label{sec:dist_estim}
Once we have frame-by-frame detection for vehicles with unique IDs, we can determine the distance each distinct vehicle has traveled on the network link. We use the timestamps for each frame as well as the GPS information of the camera-mounted vehicle, called the \textit{probe vehicle} from now on. Using frame-by-frame timestamps and the distances of all detected vehicles, we can generate a time-space diagram for the link.

We calculate the distance traveled by a detected vehicle by summing the distance of the \textit{probe vehicle} on the link and the distance of the detected vehicle from the \textit{probe vehicle}. For each timestamp $t$, the distance of the $i$ detected vehicle can be calculated as:

\begin{equation}
\label{eq:vehDist}
d^{t}_{i} = d^{t}_{probe} + d^{t}_{i, probe}
\end{equation}
where $d^{t}_{probe}$ is the distance traveled by the \textit{probe vehicle} on the link at time $t$ and $d^{t}_{i, probe}$ is the distance of the $i$ detected vehicle from the \textit{probe vehicle} at time $t$.

The value of $d^{t}_{probe}$ is derived from the metadata of the \textit{probe vehicle}. The camera used in the project is equipped with a GNSS sensor to measure the latitude and longitude of the \textit{probe vehicle} for each frame. Given the predetermined GPS location of the start of the link, we can find $d^{t}_{probe}$ by calculating the geodesic distance between the two GPS coordinates at each timestamp $t$. To calculate the distance, we use the geodesic method proposed by Charles F. Karney \cite{Karney2011AlgorithmsGeodesics}. This geodesic computation algorithm has advantages over classical algorithms, such as an always converging solution and full double precision accuracy for the solution of geodesics.

The value of $d^{t}_{i, probe}$ is calculated using a photogrammetry algorithm based on the similarity triangle. Similar to other research \cite{Kumar2021CitywideVideos, Nienaber2015AEstimation}, we use the similarity triangle algorithm to calculate the distance of $i$ detected vehicle at each frame of the video sequence.  The similarity triangle approach states that the ratio of the height of the object ${i}$ in the real world (in meters), $H^{i, Real}_{m}$ and height of the object in image dimensions (in pixels),  $H^{i, Img}_{px}$ can be described as: 

\begin{equation}
\frac{H^{i, Real}_{m}}{H^{i, Img}_{px}} = \frac{D^{i}_{m}}{F_{px}}
\label{eq:ratio}
\end{equation}
where $D^{i}_{m}$ is the distance of the object from the camera (in meters) and $F_{px}$ is the focal length of the camera (in pixels). The value of $H^{i, Img}_{px}$ can be calculated using the image height, $I_{px}$, the image sensor height, $S_{px}$, and the height of the object in the image (in pixels), $H_{i,px}$. The relationship is given by:

\begin{equation}
H^{i, Img}_{px} = \frac{S_{px}*H^{i}_{px}}{I_{px}}
\label{eq:h_in_mm}
\end{equation}

Using Eq.\ref{eq:ratio} and Eq.\ref{eq:h_in_mm} and reshuffling them, we can calculate the distance of the object from the camera (in meters), as:

\begin{equation}
D^{i}_{m} = \frac{ F_{px} * H^{i, Real}_{m} * I_{px} }{H^{i}_{px} * S_{px}}
\label{eq:dist_in_meter}
\end{equation}

For the purpose of analysis, we derive the values of parameters $H^{i, Real}_{m}$, $I_{px}$, $F_{px}$ and $S_{px}$ from the KITTI dataset. The values of these parameters used for analysis are presented in Table \ref{tbl:lane_parms}. The average height of the categories is calculated using the annotated 3D bounding box labels in the \textit{KITTI 2D box tracking} dataset. The rest of the parameters refer to the intrinsic properties of the camera used in the KITTI data collection. 

\begin{table}
\renewcommand{\arraystretch}{1.3}
\caption{Parameters used for lane distance estimation from KITTI dataset.\cite{Geiger2013VisionDataset, Geiger2012AreSuite}}
\label{tbl:lane_parms}
\centering
\begin{tabular}{|l|l|c|}
\hline
\multicolumn{1}{|c|}{\textbf{Parameter}} & \multicolumn{1}{c|}{\textbf{Explanation}}   & \textbf{Value} \\ \hline
$H^{Car, Real}$                          & Avg. height of car in KITTI         & 1.50 m            \\ \hline
$I_{px}$                                 & Image sensor height                 & 376  px          \\ \hline
$F_{px}$                                 & Focal length of camera              & 721  px              \\ \hline
$S_{px}$                                 & Image sensor height                 & 362  px           \\ \hline
\end{tabular}
\end{table}

\section{Results}\label{sec:results}
We present the results of the training of the YOLOv5 object detector on \textit{KITTI 2D object detection} as well as the evaluation of StrongSORT tracking and the lane distance estimation method on \textit{KITTI 2D box tracking}. Finally, we conclude the section with a comparison of the time-space diagram generated using the ground-truth data and the proposed methodology. 

\subsection{Object Detection Results}
The evaluation of the YOLOv5 model trained on the \textit{KITTI 2D object detection} dataset is carried out on a test set comprising 746 images. The model's performance was assessed using mean Average Precision (mAP), a standard metric for object detection benchmarks \cite{Padilla2020AAlgorithms}. In Table \ref{tbl:test_eval_yolo}, we show the average mAP value for IoU thresholds between 0.5 and 0.95 (mAP[.5 - .95]), along with the precision (P) and recall (R) for each category and the overall results.



 On the basis of the evaluation results, we observed that the trained network performs exceptionally well in detecting "car"-like objects such as "car", "van", and "truck" compared to other categories. This finding can be attributed to the relatively higher number of instances in those categories and their rigid structure. On the contrary, categories like "pedestrian" and "cyclist" exhibit more structural variations, resulting in lower prediction capabilities of the model due to the limited variation in the training dataset. In the context of the proposed methodology, it is advantageous to have high accuracy specifically in detecting the categories "car" and "van", since the focus is on identifying motorized vehicles on the road in the subsequent steps.

\begin{table}
\renewcommand{\arraystretch}{1.3}
\caption{Object Detection Evaluation Results.}
\label{tbl:test_eval_yolo}
\centering
\begin{tabular}{|c|c|c|c|c|}
\hline
{\textbf{Class}} & {\textbf{Instances}} & {\textbf{P}} & {\textbf{R}} & {\textbf{mAP{[}.5-.95{]}}} \\ \hline
Car                  & 2923                     & 0.926            & 0.933            & 0.789                          \\ \hline
Van                  & 285                      & 0.898            & 0.930             & 0.785                          \\ \hline
Truck                & 107                      & 0.943            & 0.944            & 0.784                          \\ \hline
Tram                 & 38                       & 0.933            & 0.868            & 0.729                          \\ \hline
Misc                 & 93                       & 0.929            & 0.903            & 0.660                           \\ \hline
Cyclist              & 141                      & 0.873            & 0.794            & 0.556                          \\ \hline
Pedestrian           & 427                      & 0.917            & 0.756            & 0.498                          \\ \hline
Person\_sitting      & 19                       & 0.602            & 0.797            & 0.456                          \\ \hline
\textbf{all}         & \textbf{4033}            & \textbf{0.878}   & \textbf{0.866}   & \textbf{0.657}                 \\ \hline
\end{tabular}
\end{table}

\subsection{Multi-Object Tracking Results}
The performance of the StrongSORT multi-object tracker was assessed using 21 annotated videos from the \textit{KITTI 2D box tracking} dataset. The tracker's evaluation was conducted using the Higher Order Tracking Accuracy (HOTA) metric, which measures performance in terms of detection, association, and localization \cite{Luiten2020HOTA:Tracking}.


To further evaluate its performance, we compared the results of the StrongSORT tracker with those of the Combined Image- and World-Space Tracking (CIWT) \cite{Osep2017CombinedScenes} and ByteTrack \cite{Zhang2022ByteTrack:Box} trackers. Both CIWT and ByteTrack are state-of-the-art trackers within the tracking-by-detection paradigm. For both ByteTrack and StrongSORT, we employed the same YOLOv5 model trained on the\textit{ KITTI 2D object detection} dataset as the detection model. All three trackers were specifically evaluated for the "car" category in the KITTI dataset. The HOTA metric results for all three trackers are presented in Table \ref{tbl:strongSORT_eval}.


The analysis of the HOTA results clearly indicates that StrongSORT exhibits excellent performance on the KITTI dataset, surpassing the other two trackers in all aspects, including accuracy for detection (DetA), association (AssA), and localization (LocA).

\begin{table}
\renewcommand{\arraystretch}{1.3}
\caption{Multi-Object Tracking Evaluation for category "car".}
\label{tbl:strongSORT_eval}
\centering
\begin{tabular}{|c|c|c|c|c|}
\hline
\textbf{Tracker} & \textbf{DetA} & \textbf{AssA} & \textbf{LocA} & \textbf{HOTA} \\ \hline
ByteTrack        & 44.311        & 62.176        & 80.916        & 51.933        \\ \hline
CIWT             & 63.111        & 70.562        & 81.609        & 66.306        \\ \hline
StrongSORT       & 76.904        & 72.719        & 89.221        & 74.639        \\ \hline
\end{tabular}
\end{table}

\begin{figure}
\centering
\includegraphics[width=3.4in]{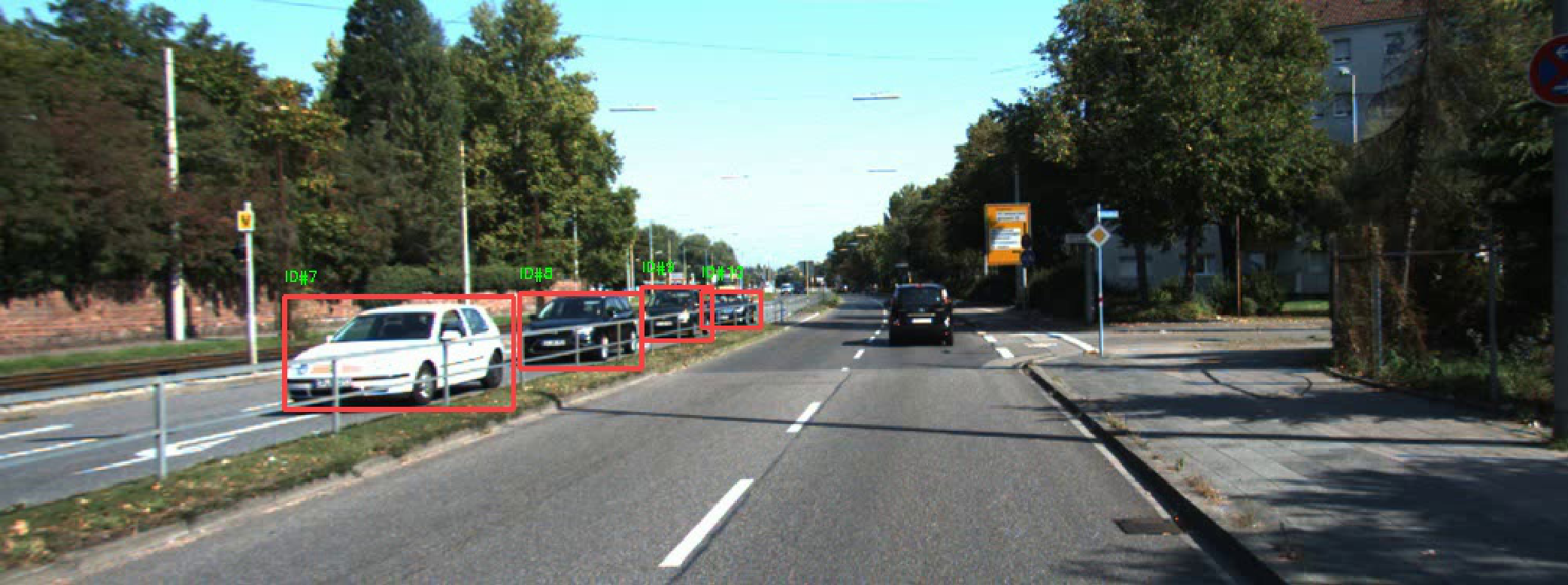}
\caption{KITTI image frame showcasing the bounding boxes (in RED) and tracking IDs (in GREEN) for vehicles in the opposite lane derived using YOLOv5m and StrongSORT.}
\label{fig:result_example}
\end{figure}

Figure \ref{fig:result_example} shows an image frame from the KITTI dataset that shows the bounding boxes and tracking IDs for each "car" found only on the opposite lane using a retrained YOLOv5m object detector and StrongSORT tracker.

\subsection{Lane Distance Estimation Results}
The accuracy of the generated time-space diagram is contingent upon the precise calculation of the distance of each vehicle detected by the camera using the similarity triangle approach. To evaluate the performance of this approach, we employ the root mean square error (RMSE) as a metric on 21 videos from the \textit{KITTI 2D box tracking} dataset.

During the evaluation, we compared the camera distance calculated using the bounding boxes for each true positive detection for "car" from the re-trained YOLOv5m model. The RMSE value was computed between the camera distance derived from the similarity triangle approach and the true depth values for each frame in each video. Additionally, we sought to ascertain the degree of error that the YOLOv5m model's bounding boxes had produced. Therefore, we also calculated the distance from the camera using the ground truth bounding boxes.

\begin{figure}
\centering
\includegraphics[width=3.2in]{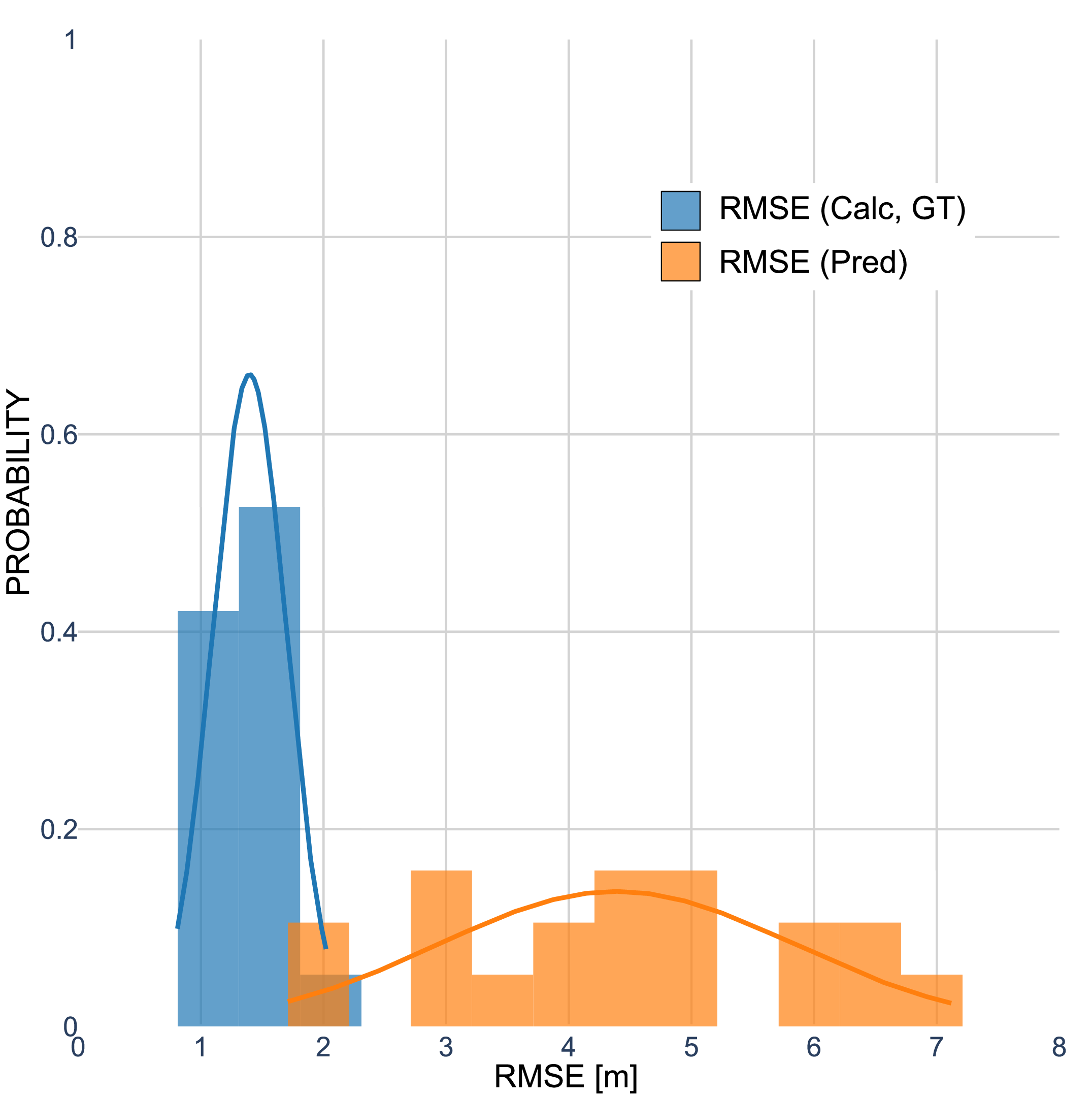}
\caption{RMSE distribution between ground truth depth and camera distance in two scenarios: RMSE (Pred) (in ORANGE) uses YOLOv5m bounding boxes for calculation, while RMSE (Calc, GT) (in BLUE) uses ground truth bounding boxes.}
\label{fig:RMSEDist}
\end{figure}

Fig. \ref{fig:RMSEDist} presents the RMSE values for a total of 24939 instances of a "car" across all videos. In the figure, \textit{"RMSE (Calc, GT)"} shows the probability distribution of RMSE calculated between the ground truth depth and the camera distance based on known bounding box values, with a mean of 1.40 [m] and a standard deviation of 0.29 [m]. In the same way, \textit{"RMSE (Pred)"} shows the probability distribution of RMSE calculated between the ground truth depth and the camera distance calculated using the bounding boxes from the re-trained YOLOv5m model, with a mean of 4.41 [m] and a standard deviation of 1.42 [m].

The results revealed that there is a considerable amount of error in calculating the distance using the proposed method. The error in the distance calculation can come from two sources. First, there is some error attributed to the limitations of the similarity triangle approach to accurately calculate the distance. It is evident from the \textit{"RMSE (Calc, GT)"} values with a high mean, pointing to a bias in the calculation. This can be because of incorrect camera intrinsic values and assumptions made about the height of the "car" in the real world. Second, inaccuracies in the retrained YOLOv5 model in determining precise bounding box values also contribute to the overall error in the distance calculation.

In order to mitigate the errors associated with the similarity triangle method, advanced algorithms based on deep learning can be integrated \cite{Lee2022VehicleDistance}, or additional data sources such as LiDAR/RADAR sensors can be utilized \cite{Ponte2017Survey}. Moreover, our observations align with previous research \cite{Nienaber2015AEstimation}, which also noted that errors tend to increase as vehicles move farther away from the camera. In their study, the authors proposed mitigation approaches to reduce such errors. These enhancements would enable more accurate distance calculations.


\subsection{Time-space Diagram Results}
Finally, a comparison is conducted between the time-space diagrams generated using ground truth data and the proposed methodology, focusing on a specific street-view video sequence. Video ID 0004 from the \textit{KITTI 2D box tracking} dataset is employed for this evaluation. The distance of each detected vehicle on the network link is calculated using the raw GPS values, following the equation described in Eq. \ref{eq:vehDist}.

\begin{figure}
\centering
\includegraphics[width=3.4in]{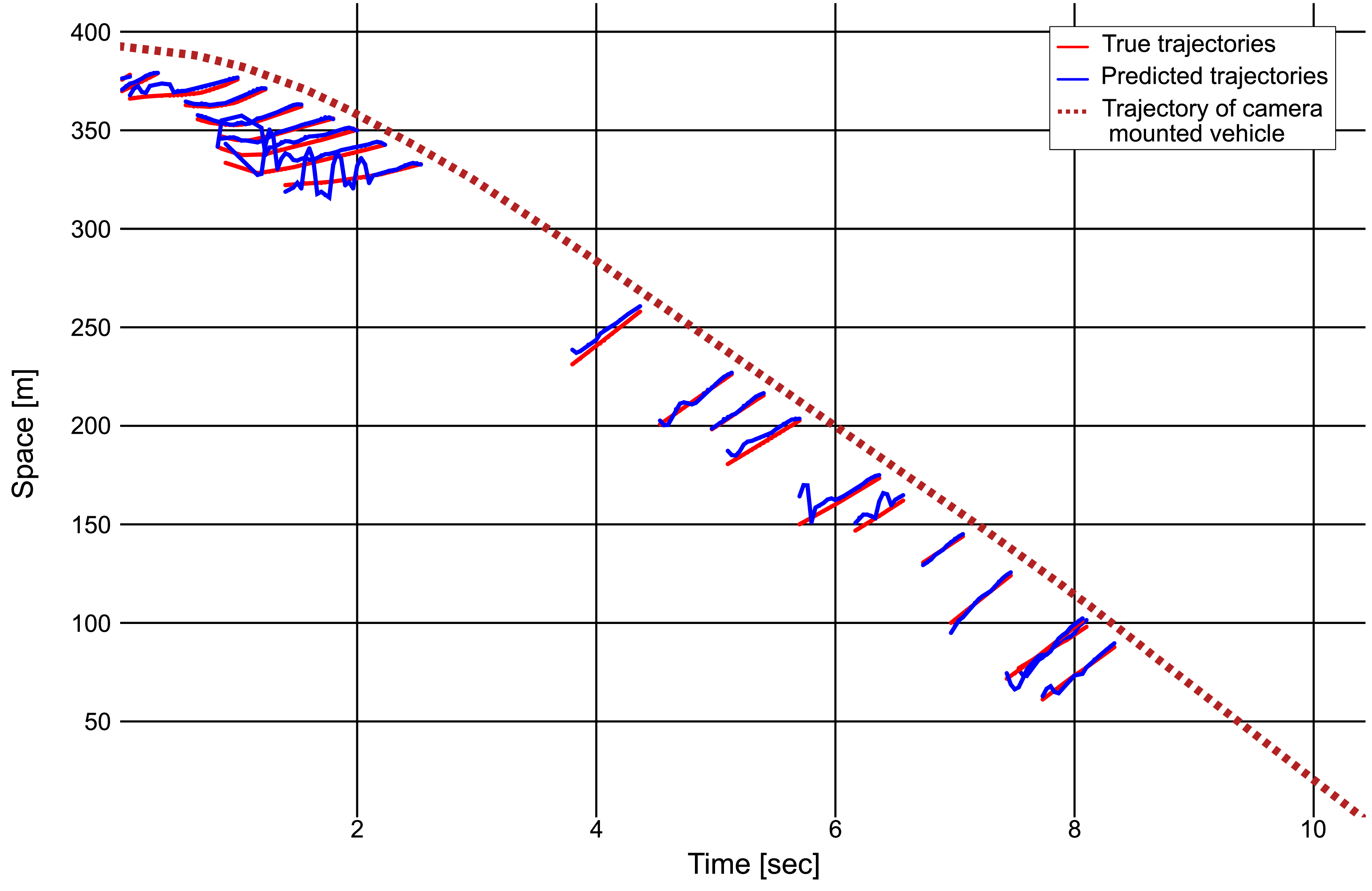}
\caption{Time-space diagram generated for video 0004.mp4 from KITTI dataset with true trajectories (in RED) and predicted trajectories (in BLUE).}
\label{fig:video_0004}
\end{figure}

Fig.~\ref{fig:video_0004} showcases the time-space diagram for video ID 0004, displaying the trajectories generated by both the actual data and our proposed methodology. It should be noted that the trajectories plotted on the time-space diagram are raw inferences from the proposed methodology and have not undergone any smoothing procedures. The RMSE value between the true trajectories and the trajectories generated by our proposed methodology exhibits a mean of 2.97 [m] and a standard deviation of 1.91 [m].

Upon observing the plot, it becomes apparent that some of the trajectories are not accurately generated using our proposed methodology. Certain errors are attributed to flickering in the bounding boxes when tracking with StrongSORT. This flickering leads to irregular bounding boxes, which subsequently generate erroneous distance values when calculated using the similarity triangle approach. Additionally, other errors discussed in the previous sections, arising from both detection and distance calculation, can also contribute to inaccuracies in trajectory inference. Some of the errors in trajectory extraction because of flickering can be reduced using the trajectory smoothing techniques highlighted in \cite{Zhang2022MonocularReview}.

\section{Conclusion} \label{sec:conclusion}
This paper introduces an innovative approach for extracting time-space diagrams of road network links from street-view video sequences captured by a camera mounted on a moving vehicle. Our proposed methodology incorporates the YOLOv5 object detector, the StrongSORT multi-object tracker, and the similarity triangle approach for distance calculation.

The evaluation results conducted on the KITTI dataset provide compelling indications that our proposed methodology is capable of generating time-space diagrams that closely resemble the actual ground truth, despite the presence of some inherent errors. These errors mainly arise from the performance of the detector and tracker utilized within the methodology. However, these issues can be mitigated through model re-training using relevant data to enhance their accuracy and robustness. Even though the simple similarity triangle method used to figure out the distance between lanes works well, it could be made better by using more complex algorithms or adding more data sources, like LIDAR, to get even more accurate distance calculations.

Advancing the research, the trajectories extracted from the time-space diagram have the potential to be utilized for estimating traffic states throughout the complete road network as well as within specific road segments. Furthermore, an evaluation of the performance of the proposed approach can encompass a range of traffic scenarios, taking into account variables such as speed, density, and volume in the opposing lane. Alternatively, there is an opportunity to explore data fusion techniques, integrating data from stationary sensors like loop detectors to enhance the analytical process.

\addtolength{\textheight}{-9cm}  

\bibliographystyle{IEEEtran}
\bibliography{IEEEabrv,references.bib}

\begin{thebibliography}{10}
\providecommand{\url}[1]{#1}
\csname url@samestyle\endcsname
\providecommand{\newblock}{\relax}
\providecommand{\bibinfo}[2]{#2}
\providecommand{\BIBentrySTDinterwordspacing}{\spaceskip=0pt\relax}
\providecommand{\BIBentryALTinterwordstretchfactor}{4}
\providecommand{\BIBentryALTinterwordspacing}{\spaceskip=\fontdimen2\font plus
\BIBentryALTinterwordstretchfactor\fontdimen3\font minus
  \fontdimen4\font\relax}
\providecommand{\BIBforeignlanguage}[2]{{%
\expandafter\ifx\csname l@#1\endcsname\relax
\typeout{** WARNING: IEEEtran.bst: No hyphenation pattern has been}%
\typeout{** loaded for the language `#1'. Using the pattern for}%
\typeout{** the default language instead.}%
\else
\language=\csname l@#1\endcsname
\fi
#2}}
\providecommand{\BIBdecl}{\relax}
\BIBdecl

\bibitem{Wang2020ATrajectories}
\BIBentryALTinterwordspacing
P.~Wang, P.~Li, F.~Chowdhury, L.~Zhang, and X.~Zhou, ``{A mixed integer
  programming formulation and scalable solution algorithms for traffic control
  coordination across multiple intersections based on vehicle space-time
  trajectories},'' \emph{Transportation Research Part B: Methodological}, vol.
  134, pp. 266--304, 4 2020. [Online]. Available:
  \url{https://linkinghub.elsevier.com/retrieve/pii/S0191261519303844}
\BIBentrySTDinterwordspacing

\bibitem{Peng2023Network-WideTrajectories}
\BIBentryALTinterwordspacing
X.~Peng and H.~Wang, ``{Network-Wide Coordinated Control Based on Space-Time
  Trajectories},'' \emph{IEEE Intelligent Transportation Systems Magazine}, pp.
  2--16, 2023. [Online]. Available:
  \url{https://ieeexplore.ieee.org/document/10064357/}
\BIBentrySTDinterwordspacing

\bibitem{Essa2018TrafficLevel}
\BIBentryALTinterwordspacing
M.~Essa and T.~Sayed, ``{Traffic conflict models to evaluate the safety of
  signalized intersections at the cycle level},'' \emph{Transportation Research
  Part C: Emerging Technologies}, vol.~89, pp. 289--302, 4 2018. [Online].
  Available:
  \url{https://linkinghub.elsevier.com/retrieve/pii/S0968090X18302249}
\BIBentrySTDinterwordspacing

\bibitem{Xing2022AFlow}
\BIBentryALTinterwordspacing
L.~Xing and W.~Liu, ``{A Data Fusion Powered Bi-Directional Long Short Term
  Memory Model for Predicting Multi-Lane Short Term Traffic Flow},'' \emph{IEEE
  Transactions on Intelligent Transportation Systems}, vol.~23, no.~9, pp.
  16\,810--16\,819, 9 2022. [Online]. Available:
  \url{https://ieeexplore.ieee.org/document/9492911/}
\BIBentrySTDinterwordspacing

\bibitem{KhajehHosseini2023TowardsNetwork}
\BIBentryALTinterwordspacing
M.~Khajeh~Hosseini and A.~Talebpour, ``{Towards Predicting Traffic Shockwave
  Formation and Propagation: A Convolutional Encoder–Decoder Network},''
  \emph{Journal of Transportation Engineering, Part A: Systems}, vol. 149,
  no.~4, 4 2023. [Online]. Available:
  \url{https://ascelibrary.org/doi/10.1061/JTEPBS.TEENG-7209}
\BIBentrySTDinterwordspacing

\bibitem{Nantes2016Real-timeData}
\BIBentryALTinterwordspacing
A.~Nantes, D.~Ngoduy, A.~Bhaskar, M.~Miska, and E.~Chung, ``{Real-time traffic
  state estimation in urban corridors from heterogeneous data},''
  \emph{Transportation Research Part C: Emerging Technologies}, vol.~66, pp.
  99--118, 5 2016. [Online]. Available:
  \url{https://linkinghub.elsevier.com/retrieve/pii/S0968090X15002454}
\BIBentrySTDinterwordspacing

\bibitem{Thodi2021LearningVisualizations}
\BIBentryALTinterwordspacing
B.~T. Thodi, Z.~S. Khan, S.~E. Jabari, and M.~Menendez, ``{Learning Traffic
  Speed Dynamics from Visualizations},'' \emph{2021 IEEE International
  Intelligent Transportation Systems Conference (ITSC)}, vol. 2021-September,
  pp. 1239--1244, 5 2021. [Online]. Available:
  \url{http://arxiv.org/abs/2105.01423
  http://dx.doi.org/10.1109/ITSC48978.2021.9564541}
\BIBentrySTDinterwordspacing

\bibitem{Zang2019ExperimentalBeijing}
\BIBentryALTinterwordspacing
J.~rui Zang, G.~hua Song, R.~ti~E, J.~ping Sun, X.~Zhang, and L.~Yu,
  ``Experimental findings about wide moving jams: Case study in beijing,''
  \emph{Journal of Transportation Engineering, Part A: Systems}, vol. 145, 7
  2019. [Online]. Available:
  \url{https://ascelibrary.org/doi/10.1061/JTEPBS.0000250}
\BIBentrySTDinterwordspacing

\bibitem{Li2017AnalysisArrivals}
\BIBentryALTinterwordspacing
C.-Y. Li, H.-J. Huang, and T.-Q. Tang, ``Analysis of user equilibrium for
  staggered shifts in a single-entry traffic corridor with no late arrivals,''
  \emph{Physica A: Statistical Mechanics and its Applications}, vol. 474, pp.
  8--18, 5 2017. [Online]. Available:
  \url{https://linkinghub.elsevier.com/retrieve/pii/S0378437117300729}
\BIBentrySTDinterwordspacing

\bibitem{Zhang2022MonocularReview}
\BIBentryALTinterwordspacing
X.~Zhang, Y.~Feng, P.~Angeloudis, and Y.~Demiris, ``{Monocular Visual Traffic
  Surveillance: A Review},'' \emph{IEEE Transactions on Intelligent
  Transportation Systems}, vol.~23, no.~9, pp. 14\,148--14\,165, 9 2022.
  [Online]. Available: \url{https://ieeexplore.ieee.org/document/9714212/}
\BIBentrySTDinterwordspacing

\bibitem{Sunderrajan2016TrafficData}
\BIBentryALTinterwordspacing
A.~Sunderrajan, V.~Viswanathan, W.~Cai, and A.~Knoll, ``{Traffic state
  estimation using floating car data},'' in \emph{Procedia Computer Science},
  vol.~80.\hskip 1em plus 0.5em minus 0.4em\relax Elsevier B.V., 2016, pp.
  2008--2018. [Online]. Available:
  \url{https://linkinghub.elsevier.com/retrieve/pii/S1877050916310110}
\BIBentrySTDinterwordspacing

\bibitem{Yuan2021TrafficData}
\BIBentryALTinterwordspacing
Y.~Yuan, W.~Zhang, X.~Yang, Y.~Liu, Z.~Liu, and W.~Wang, ``Traffic state
  classification and prediction based on trajectory data,'' \emph{Journal of
  Intelligent Transportation Systems}, pp. 1--15, 9 2021. [Online]. Available:
  \url{https://www.tandfonline.com/doi/full/10.1080/15472450.2021.1955210}
\BIBentrySTDinterwordspacing

\bibitem{Rahmani2010RequirementsStudy}
\BIBentryALTinterwordspacing
M.~Rahmani, H.~N. Koutsopoulos, and A.~Ranganathan, ``Requirements and
  potential of gps-based floating car data for traffic management: Stockholm
  case study,'' \emph{13th International IEEE Conference on Intelligent
  Transportation Systems}, pp. 730--735, 9 2010. [Online]. Available:
  \url{http://ieeexplore.ieee.org/document/5625177/}
\BIBentrySTDinterwordspacing

\bibitem{Ua-Areemitr2019Low-CostProcessing}
\BIBentryALTinterwordspacing
E.~Ua-areemitr, A.~Sumalee, and W.~H. Lam, ``Low-cost road traffic state
  estimation system using time-spatial image processing,'' \emph{IEEE
  Intelligent Transportation Systems Magazine}, vol.~11, pp. 69--79, 9 2019.
  [Online]. Available: \url{https://ieeexplore.ieee.org/document/8742556/}
\BIBentrySTDinterwordspacing

\bibitem{Pletzer2012RobustCameras}
\BIBentryALTinterwordspacing
F.~Pletzer, R.~Tusch, L.~Boszormenyi, and B.~Rinner, ``{Robust Traffic State
  Estimation on Smart Cameras},'' in \emph{2012 IEEE Ninth International
  Conference on Advanced Video and Signal-Based Surveillance}.\hskip 1em plus
  0.5em minus 0.4em\relax IEEE, 9 2012, pp. 434--439. [Online]. Available:
  \url{http://ieeexplore.ieee.org/document/6328053/}
\BIBentrySTDinterwordspacing

\bibitem{Seo2015EstimationEquipment}
\BIBentryALTinterwordspacing
T.~Seo, T.~Kusakabe, and Y.~Asakura, ``Estimation of flow and density using
  probe vehicles with spacing measurement equipment,'' \emph{Transportation
  Research Part C: Emerging Technologies}, vol.~53, pp. 134--150, 2015.
  [Online]. Available:
  \url{https://www.sciencedirect.com/science/article/pii/S0968090X15000443}
\BIBentrySTDinterwordspacing

\bibitem{Cao2011MobileIntegration}
\BIBentryALTinterwordspacing
M.~Cao, W.~Zhu, and M.~Barth, ``Mobile traffic surveillance system for dynamic
  roadway and vehicle traffic data integration,'' \emph{2011 14th International
  IEEE Conference on Intelligent Transportation Systems (ITSC)}, pp. 771--776,
  10 2011. [Online]. Available:
  \url{http://ieeexplore.ieee.org/document/6083096/}
\BIBentrySTDinterwordspacing

\bibitem{Guerrieri2021DeepTechnique}
\BIBentryALTinterwordspacing
M.~Guerrieri and G.~Parla, ``{Deep Learning and YOLOv3 Systems for Automatic
  Traffic Data Measurement by Moving Car Observer Technique},''
  \emph{Infrastructures}, vol.~6, no.~9, p. 134, 9 2021. [Online]. Available:
  \url{https://www.mdpi.com/2412-3811/6/9/134}
\BIBentrySTDinterwordspacing

\bibitem{Kumar2021CitywideVideos}
\BIBentryALTinterwordspacing
A.~Kumar, T.~Kashiyama, H.~Maeda, H.~Omata, and Y.~Sekimoto, ``Real-time
  citywide reconstruction of traffic flow from moving cameras on lightweight
  edge devices,'' \emph{ISPRS Journal of Photogrammetry and Remote Sensing},
  vol. 192, pp. 115--129, 2022. [Online]. Available:
  \url{https://www.sciencedirect.com/science/article/pii/S092427162200199X}
\BIBentrySTDinterwordspacing

\bibitem{Kumar2022CARLA}
------, ``Citywide reconstruction of traffic flow using the vehicle-mounted
  moving camera in the carla driving simulator,'' in \emph{2022 IEEE 25th
  International Conference on Intelligent Transportation Systems (ITSC)}, 2022,
  pp. 2292--2299.

\bibitem{Geiger2012AreSuite}
\BIBentryALTinterwordspacing
A.~Geiger, P.~Lenz, and R.~Urtasun, ``{Are we ready for autonomous driving? the
  KITTI vision benchmark suite},'' in \emph{Proceedings of the IEEE Computer
  Society Conference on Computer Vision and Pattern Recognition}, 2012, pp.
  3354--3361. [Online]. Available:
  \url{https://ieeexplore.ieee.org/document/6248074}
\BIBentrySTDinterwordspacing

\bibitem{Luiten2020TrackEval}
\BIBentryALTinterwordspacing
J.~Luiten and A.~Hoffhues, ``{TrackEval},'' 2020. [Online]. Available:
  \url{https://github.com/JonathonLuiten/TrackEval}
\BIBentrySTDinterwordspacing

\bibitem{Nepal2022ComparingUAVs.}
\BIBentryALTinterwordspacing
U.~Nepal and H.~Eslamiat, ``{Comparing YOLOv3, YOLOv4 and YOLOv5 for Autonomous
  Landing Spot Detection in Faulty UAVs.}'' \emph{Sensors (Basel,
  Switzerland)}, vol.~22, no.~2, 1 2022. [Online]. Available:
  \url{http://www.ncbi.nlm.nih.gov/pubmed/35062425}
\BIBentrySTDinterwordspacing

\bibitem{GeYOLOX:Detectors}
\BIBentryALTinterwordspacing
Z.~Ge, S.~Liu, F.~Wang, Z.~Li, and J.~Sun, ``Yolox: Exceeding yolo series in
  2021,'' p.~12, 7 2021. [Online]. Available:
  \url{http://arxiv.org/abs/2107.08430}
\BIBentrySTDinterwordspacing

\bibitem{GlennJocher2022YOLOv5}
\BIBentryALTinterwordspacing
{Glenn Jocher}, ``{YOLOv5},'' 2022. [Online]. Available:
  \url{https://github.com/ultralytics/yolov5}
\BIBentrySTDinterwordspacing

\bibitem{Lin2014MicrosoftContext}
\BIBentryALTinterwordspacing
T.-Y. Lin, M.~Maire, S.~Belongie, L.~Bourdev, R.~Girshick, J.~Hays, P.~Perona,
  D.~Ramanan, C.~L. Zitnick, and P.~Dollár, ``Microsoft coco: Common objects
  in context,'' 5 2014. [Online]. Available:
  \url{http://arxiv.org/abs/1405.0312}
\BIBentrySTDinterwordspacing

\bibitem{Elfwing2018Sigmoid-weightedLearning}
\BIBentryALTinterwordspacing
S.~Elfwing, E.~Uchibe, and K.~Doya, ``{Sigmoid-weighted linear units for neural
  network function approximation in reinforcement learning},'' \emph{Neural
  Networks}, vol. 107, pp. 3--11, 11 2018. [Online]. Available:
  \url{https://linkinghub.elsevier.com/retrieve/pii/S0893608017302976}
\BIBentrySTDinterwordspacing

\bibitem{Du2022StrongSORT:Again}
\BIBentryALTinterwordspacing
Y.~Du, Z.~Zhao, Y.~Song, Y.~Zhao, F.~Su, T.~Gong, and H.~Meng, ``Strongsort:
  Make deepsort great again,'' 2 2022. [Online]. Available:
  \url{http://arxiv.org/abs/2202.13514}
\BIBentrySTDinterwordspacing

\bibitem{Brostrom2022RealtimeOSNet}
\BIBentryALTinterwordspacing
{Mikel Broström}, ``{Real-time multi-camera multi-object tracker using YOLOv7
  and StrongSORT with OSNet},'' 2022. [Online]. Available:
  \url{https://github.com/mikel-brostrom/Yolov7\_StrongSORT\_OSNet}
\BIBentrySTDinterwordspacing

\bibitem{Karney2011AlgorithmsGeodesics}
\BIBentryALTinterwordspacing
C.~F.~F. Karney, ``{Algorithms for geodesics},'' \emph{Journal of Geodesy},
  vol.~87, no.~1, pp. 43--55, 9 2011. [Online]. Available:
  \url{http://arxiv.org/abs/1109.4448
  http://dx.doi.org/10.1007/s00190-012-0578-z}
\BIBentrySTDinterwordspacing

\bibitem{Nienaber2015AEstimation}
\BIBentryALTinterwordspacing
S.~Nienaber, R.~Kroon, and M.~Booysen, ``{A Comparison of Low-Cost Monocular
  Vision Techniques for Pothole Distance Estimation},'' in \emph{2015 IEEE
  Symposium Series on Computational Intelligence}.\hskip 1em plus 0.5em minus
  0.4em\relax IEEE, 12 2015, pp. 419--426. [Online]. Available:
  \url{http://ieeexplore.ieee.org/document/7376642/}
\BIBentrySTDinterwordspacing

\bibitem{Geiger2013VisionDataset}
\BIBentryALTinterwordspacing
A.~Geiger, P.~Lenz, C.~Stiller, and R.~Urtasun, ``{Vision meets robotics: The
  KITTI dataset},'' \emph{The International Journal of Robotics Research},
  vol.~32, no.~11, pp. 1231--1237, 9 2013. [Online]. Available:
  \url{http://journals.sagepub.com/doi/10.1177/0278364913491297}
\BIBentrySTDinterwordspacing

\bibitem{Padilla2020AAlgorithms}
\BIBentryALTinterwordspacing
R.~Padilla, S.~L. Netto, and E.~A.~B. da~Silva, ``{A Survey on Performance
  Metrics for Object-Detection Algorithms},'' in \emph{2020 International
  Conference on Systems, Signals and Image Processing (IWSSIP)}.\hskip 1em plus
  0.5em minus 0.4em\relax IEEE, 7 2020, pp. 237--242. [Online]. Available:
  \url{https://ieeexplore.ieee.org/document/9145130/}
\BIBentrySTDinterwordspacing

\bibitem{Luiten2020HOTA:Tracking}
\BIBentryALTinterwordspacing
J.~Luiten, A.~Osep, P.~Dendorfer, P.~Torr, A.~Geiger, L.~Leal-Taixe, and
  B.~Leibe, ``{HOTA: A Higher Order Metric for Evaluating Multi-Object
  Tracking},'' \emph{International Journal of Computer Vision}, 9 2020.
  [Online]. Available: \url{http://arxiv.org/abs/2009.07736
  http://dx.doi.org/10.1007/s11263-020-01375-2}
\BIBentrySTDinterwordspacing

\bibitem{Osep2017CombinedScenes}
\BIBentryALTinterwordspacing
A.~Osep, W.~Mehner, M.~Mathias, and B.~Leibe, ``{Combined image- and
  world-space tracking in traffic scenes},'' in \emph{2017 IEEE International
  Conference on Robotics and Automation (ICRA)}.\hskip 1em plus 0.5em minus
  0.4em\relax IEEE, 5 2017, pp. 1988--1995. [Online]. Available:
  \url{http://ieeexplore.ieee.org/document/7989230/}
\BIBentrySTDinterwordspacing

\bibitem{Zhang2022ByteTrack:Box}
\BIBentryALTinterwordspacing
Y.~Zhang, P.~Sun, Y.~Jiang, D.~Yu, F.~Weng, Z.~Yuan, P.~Luo, W.~Liu, and
  X.~Wang, ``{ByteTrack: Multi-object Tracking by Associating Every Detection
  Box},'' in \emph{Computer Vision -- ECCV 2022}, ser. Lecture Notes in
  Computer Science, S.~Avidan, G.~Brostow, M.~Ciss{\'{e}}, G.~M. Farinella, and
  T.~Hassner, Eds.\hskip 1em plus 0.5em minus 0.4em\relax Cham: Springer Nature
  Switzerland, 2022, vol. 13682, pp. 1--21. [Online]. Available:
  \url{https://link.springer.com/10.1007/978-3-031-20047-2\_1}
\BIBentrySTDinterwordspacing

\bibitem{Lee2022VehicleDistance}
\BIBentryALTinterwordspacing
S.~Lee, K.~Han, S.~Park, and X.~Yang, ``Vehicle distance estimation from a
  monocular camera for advanced driver assistance systems,'' \emph{Symmetry},
  vol.~14, p. 2657, 12 2022. [Online]. Available:
  \url{https://www.mdpi.com/2073-8994/14/12/2657}
\BIBentrySTDinterwordspacing

\bibitem{Ponte2017Survey}
\BIBentryALTinterwordspacing
F.~de~Ponte~Müller, ``Survey on ranging sensors and cooperative techniques for
  relative positioning of vehicles,'' \emph{Sensors}, vol.~17, p. 271, 1 2017.
  [Online]. Available: \url{http://www.mdpi.com/1424-8220/17/2/271}
\BIBentrySTDinterwordspacing

\end{thebibliography}

\end{document}